\documentclass[letterpaper, 10 pt, conference]{ieeeconf}  

\IEEEoverridecommandlockouts                              

\usepackage{graphicx}
\usepackage[outdir=./]{epstopdf}
\DeclareGraphicsExtensions{.eps,.jpg}
\usepackage[table]{xcolor}
\usepackage{balance}
\usepackage{flushend}

\usepackage{siunitx}
\usepackage{amsmath}
\usepackage{caption}
\usepackage{subcaption}
\usepackage{tablefootnote}

\usepackage{multirow}

\usepackage{amssymb}
\usepackage{algorithm}

\usepackage{hyperref}
\usepackage{tikz}
\usepackage{algpseudocode}
\usepackage{xspace}
\usepackage{tabularx} 
\usepackage{array}
\usepackage[hang,flushmargin]{footmisc}

\newcommand\tab[1][0.5cm]{\hspace*{#1}}
\usepackage{accents}  

\def \poses{X}
\def \landmarks{L}
\def \associations{D}
\def \measurements{Z}

\def \association{d}

\title{\LARGE \bf
Opti-Acoustic Semantic SLAM with Unknown Objects in Underwater Environments 
}

\author{Kurran Singh$^1$, Jungseok Hong$^1$, Nicholas R. Rypkema$^2$ and John J. Leonard$^1$
\thanks{$^1$K. Singh, J. Hong, and J. Leonard are with the Computer Science and Artificial Intelligence Laboratory (CSAIL) at the Massachusetts Institute of Technology (MIT), 32 Vassar St, Cambridge, MA 02139, USA. 
$^2$ N. Rypkema is with the Applied Ocean Physics and Engineering Department at the Woods Hole Oceanographic Institute (WHOI), 266 Woods Hole Rd. Woods Hole, MA 02543, USA. 
        Corresponding author: Kurran Singh ({\tt\small singhk@mit.edu})}%
}

\begin{document}
\maketitle
\thispagestyle{empty}
\pagestyle{empty}

\begin{abstract}
Despite recent advances in semantic Simultaneous Localization and Mapping (SLAM) for terrestrial and aerial applications, underwater semantic SLAM remains an open and largely unaddressed research problem due to the unique sensing modalities and the object classes found underwater. This paper presents an object-based semantic SLAM method for underwater environments that can identify, localize, classify, and map a wide variety of marine objects without \textit{a priori} knowledge of the object classes present in the scene.
The method performs unsupervised object segmentation and object-level feature aggregation, and then uses opti-acoustic sensor fusion for object localization. Probabilistic data association is used to determine observation to landmark correspondences. Given such correspondences, the method then jointly optimizes landmark and vehicle position estimates. 
Indoor and outdoor underwater datasets with a wide variety of objects and challenging acoustic and lighting conditions are collected for evaluation and made publicly available. Quantitative and qualitative results show the proposed method achieves reduced trajectory error compared to baseline methods, and is able to obtain comparable map accuracy to a baseline closed-set method that requires hand-labeled data of all objects in the scene.
\end{abstract}
\section*{Supplementary Material}
\label{sect:supplemental-material}
Data is made available here: https://kurransingh.github.io/\\underwater-semantic-slam/
\section{Introduction}

Recent advances in state estimation and perception methods have significantly improved the capability of terrestrial and aerial robots to map environments using semantically meaningful objects~\cite{Rosinol2021},~\cite{8575532},~\cite{schmid2024khronos}. Robust and reliable semantic mapping systems are crucial for robots to achieve a higher level of autonomy, which is critical for underwater robotics applications given their often long-term and communications-limited nature. Semantic maps can also serve as a compressed map representation for multi-vehicle scenarios \cite{jamieson2021multi}, and for diver-AUV collaboration \cite{oneillthesis}. However, underwater robotics has not kept pace with terrestrial and aerial robotics in semantic mapping due to the unique challenges posed by underwater environments.

One of the primary hurdles in underwater semantic Simultaneous Localization and Mapping (SLAM) is the lack of GPS availability and the substantial attenuation of electromagnetic waves underwater. 
Moreover, underwater visibility is degraded by light scattering and color absorption, worsening perception challenges.
\begin{figure}  
    \vspace{2mm}
    \centering
    \scalebox{0.77}{\input{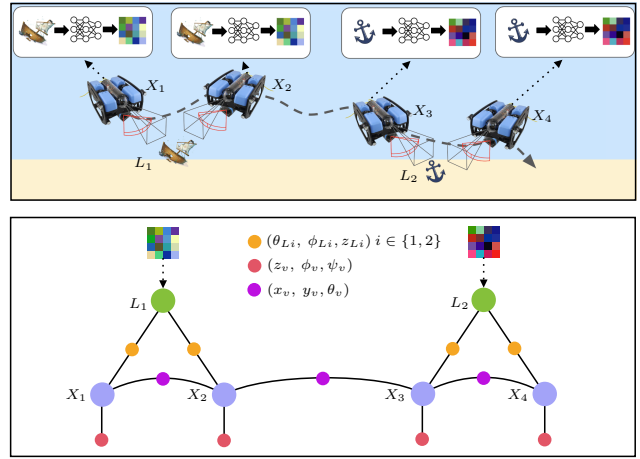}}
    \caption{An example scenario, where an underwater vehicle encounters objects during its navigation, demonstrating the proposed semantic SLAM approach. (Top): The vehicle senses objects using both optical and acoustic sensors. Optical sensing information is used to extract semantic feature embeddings of each object, and the ranges to the objects are obtained from the acoustic sensor. (Bottom): Information obtained from objects and odometry is used for building a factor graph ($X_i$: vehicle states; $L_i$: landmark states; $\theta_{Li}, \phi_{Li}, z_{Li}$: bearing, elevation, and ranges to objects; $z_{v}, \phi_v, \psi_v$: depth, pitch, and roll of the vehicle; $x_v, y_v, \theta_v$: location and yaw of the vehicle). }  
    \label{fig:intro}
    \vspace{-0mm}
\end{figure}

\begin{figure*}  
    \vspace{2mm}
    \centering
    \scalebox{0.85}{\input{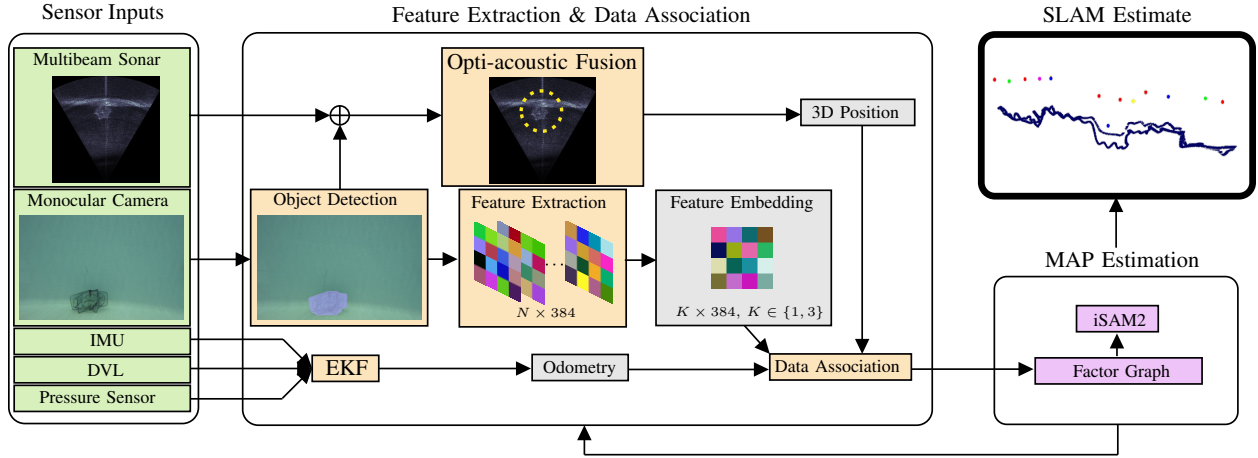}}
    \caption{The pipeline receives sensor inputs from a multibeam sonar, a monocular camera, an IMU, a DVL, and a pressure sensor. Given those sensor inputs, the system does the following: (1) Segment objects from optical images and extract features for each segmentation mask. 
    The dimension of these extracted features is dependent on the corresponding object's size in the image. 
    These features are projected into a fixed dimension for the data association process. (2) Given the segmentation mask, its pixel centroid is used to find a corresponding range from sonar returns and estimate a 3D position of the object. (3) In parallel, sensor readings from IMU, DVL, and pressure sensor are used to estimate odometry. Outputs from (1)-(3) are used to build a factor graphwhich is optimized via iSAM2 to produce maximum a posteriori (MAP) estimates. Lastly, we obtain map and trajectory estimates as an output.}  
    \label{fig:pipeline}
    \vspace{-7mm}
\end{figure*}

In an effort to tackle underwater SLAM, existing works primarily use sonar and camera sensors (\ref{subsec:optiacoustic_relatedworks}). However, sonar data is subject to multi-path reflections~\cite{Teixeira2016} and other noise-inducing phenomena; underwater camera data contains difficult lighting effects such as caustics~\cite{Lanthier2004}. 
The scarcity of large labeled datasets for both sonar and camera has made the problem more challenging. These limitations have restricted previous underwater semantic SLAM techniques to use a small number of object classes or be developed in simulation. 
Furthermore, the prohibitive cost of underwater motion capture systems has forced most existing semantic mapping methods to rely on manual measurements and image registration to estimate real-world ground truth trajectories, which are prone to error.

Addressing these challenges, we propose a novel approach to underwater semantic SLAM. 
This approach involves exploiting unsupervised object segmentation to identify objects and generate corresponding masks for accurate localization of the objects. 
Figure~\ref{fig:intro} illustrates how camera and sonar sensors are co-located for fusing each mask with sonar measurements to estimate the 3D position of the object. 
From segmented masks, feature embeddings of a fixed dimension are produced. Additionally, odometry is estimated by combining an inertial measurement unit (IMU), Doppler velocity log (DVL), and pressure sensor measurements with an Extended Kalman Filter (EKF). By incorporating these measurements into a factor graph framework, as shown in Figure~\ref{fig:intro}, the proposed method detects semantically meaningful objects, uses them as landmarks, and outputs a semantic map, while also reducing the odometric drift of the trajectories. For evaluation, the data from a low-cost Hybrid Long/Inverted Ultra-Short Baseline (iLBL-USBL) acoustic beacon system is used to obtain a ground truth trajectory. This allows for the evaluation of the method quantitatively on both mapping and trajectory accuracy. Qualitative results from various underwater environment scenarios are also presented.


Results demonstrate that the proposed approach enables an underwater vehicle to map underwater environments accurately by identifying and locating semantically meaningful objects. 
Through this method, the challenges of underwater semantic mapping including the lack of training data for object recognition and the limited visibility for stereo vision are addressed, thereby enhancing autonomy for sophisticated underwater operations.
The contributions of this work are as follows:
\begin{itemize}
    \item A novel underwater semantic SLAM system capable of mapping a wide variety of object classes without additional training, and without any prior knowledge of the scene's object classes.
    \item An opti-acoustic data fusion method that provides accurate object localization in 3D.
    \item Quantitative and qualitative evaluations of the system using real underwater data, which substantiates the map accuracy and trajectory optimization capabilities of the proposed approach.
    \item A public dataset consisting of camera, sonar, odometry, and ground truth trajectory data, as well as 691 labeled images for common underwater objects across 17 different classes. 
\end{itemize}

\section{Related Work}
This work ties together concepts from machine learning, sensor fusion, and underwater mapping. We provide a brief overview of relevant related works from the above areas. 

\subsection{Object Segmentation}
Recent advances in the Transformer architecture~\cite{attention} have inspired the development of Vision Transformers (ViTs)~\cite{vit}, which outperform convolutional neural networks (CNNs) for multiple computer vision tasks, including object segmentation, detection, and classification. However, understanding the visual representation of an image has remained extremely challenging for a model when the image was not previously seen by the model. To address this issue, Caron~\textit{et al.} proposed DINO~\cite{dino}. It exploits unsupervised learning approaches (\textit{e.g,} a teacher-student network) to distill information from large datasets without any labels using ViTs. Radford~\textit{et al.}~\cite{clip} presented multi-modal learning, which uses image-text pairs with a contrastive loss. The model, CLIP, learns how to understand images in the context of natural languages. Kirillov~\textit{et al.}~\cite{sam} proposed a promptable segmentation method segment anything (SAM). It uses a pre-trained encoder from CLIP and takes boxes, masks, and points as prompts. These models' ability to extract contextual information from images has led to the development of several segmentation models. For instance,~\cite{HSG, u2seg, cutler} proposed unsupervised segmentation models without using any explicit labels. ~\cite{groundedsam, semanticsam, sam_hq} are the examples of SAM variants. 
CLIP- and SAM-based approaches achieve comparable segmentation accuracy to fully supervised models (\textit{e.g.}, ~\cite{sam}) for terrestrial images. 

\subsection{Opti-Acoustic Correspondences}\label{subsec:optiacoustic_relatedworks}
Purely sonar-based underwater mapping has a relatively large body of work such as Teixeira~\textit{et al.}~\cite{Teixeira2016} which proposes a submap-based SLAM method for sonar mapping, and Westman~\textit{et al.}~\cite{westman2020volumetric} which describes a method for dense sonar-based reconstruction. Purely optical camera-based underwater mapping has also been explored in depth by works such as \cite{Kim2013} which performs visual SLAM with a bag-of-words saliency criteria for keyframe selection. 

Unfortunately, each sensor modality has a degeneracy stemming from the loss of the elevation angle for sonar, and the loss of the range measurement for monocular cameras. Some works seek to address the degeneracy directly for sonar-only mapping by learning the elevation angle \cite{Wang2021}, through the fusion of orthogonally mounted sonars \cite{McConnell2020}, or through carefully handling the degeneracy in the back-end \cite{westman2019degeneracy}. Stereo or multi-view for optical camera-only mapping has been explored as well, but suffers from issues with lighting conditions, reflections, caustics, and featureless areas such as plain walls \cite{Lanthier2004}. However, combining the two sensor modalities to address the degeneracies of each sensor modality is likely the best way to fully utilize their complimentary geometries. 
\begin{figure}  
    \vspace{2mm}
    \centering
    \scalebox{0.99}{\input{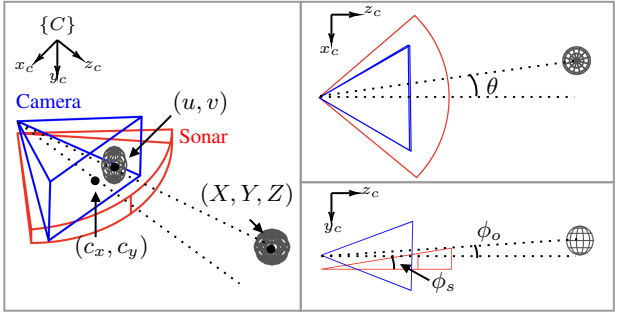}}
    \caption{Schematic representation of sensor placements of a camera and sonar. The diagram on the left illustrates the spatial configuration of a camera and sonar sensor with their respective coordinate frames, where {\(C\)} represents a camera frame. The camera's field of view is depicted in blue, and the sonar's conical beam is represented in red. The diagram on the right details the observed object's orientation angles of \(\theta, \phi_o\) from the camera and \(\phi_s\) from the sonar, depicting how angular measurements are derived.}
  
    \label{fig:opti-acoustic}
\end{figure}

Several previous works exist proposing different methods and use cases of opti-acoustic fusion, with Ferreira~\textit{et al.}~\cite{Ferreira2016} providing a survey of such opti-acoustic fusion methods. Negahdaripour~\cite{Negahdaripour2007} describes the epipolar geometry for a sonar and camera configuration, and also provides a method for opti-acoustic calibration and dense object reconstruction in \cite{Negahdaripour2009}, while Pecheux~\textit{et al.}~\cite{Pecheux2023} refine the opti-acoustic calibration process through feature matching which negates the need for a specific calibration structure. 

Jang~\textit{et al.}~\cite{jang2020cnn} attempt to learn opti-acoustic feature correspondences through the use of a convolutional neural network, while Jang~\textit{et al.}~\cite{Jang2021} try to calculate opti-acoustic feature matching by applying a style-transfer method and through integration with SuperGlue for multi-session or multi-vehicle mapping. Lanthier~\textit{et al.}~\cite{Lanthier2004} combine a stereo camera with sonar and infrared sensors to perform mapping, but note that stereo cameras struggle with obtaining range measurements for underwater vision due to uniform and featureless regions, and difficult lighting effects such as reflections and caustics. Yang~\textit{et al.}~\cite{Yang2022} use sonar to provide an absolute scale to monocular odometry. Kim~\textit{et al.}~\cite{Kim2019} separately perform visual SLAM and sonar space carving in two different mapping modules, before combining the two for dense reconstruction. Rahman~\textit{et al.}~\cite{rahman2022} combine sonar and monocular camera data in a SLAM framework they refer to as SVIN2, but their mapping technique uses point features rather than objects.

The only work that performs fusion at the \textit{object} level as required by our method, rather than at the \textit{feature} level, is by Raaj~\textit{et al.}~\cite{Raaj2016a}, who use a monocular camera and sonar in a particle filter, albeit only for a single object.

\subsection{Semantic SLAM}
While terrestrial semantic SLAM has been thoroughly explored with applications in, for example, dynamic environments (Cui~\textit{et al.}~\cite{cui2019sof}), monocular cameras (Civera~\textit{et al.}~\cite{civera2011towards}), lidar (Chen~\textit{et al.}~\cite{chen2019suma++}), probabilistic data association (Bowman~\textit{et al.}~\cite{bowman2017probabilistic}), and scene graphs (Rosinol~\textit{et al.}~\cite{Rosinol2021}), far fewer such works exist for underwater scenarios. 
Arain~\textit{et al.}~\cite{Arain19} use a stereo camera to perform binary classifications on objects for obstacle avoidance. Himri~\textit{et al.}~\cite{HIMRI2018360} and Vallicrosa~\textit{et al.}~\cite{s21206740} rely on a database of a priori known objects, and use a laser scanner to detect objects from the database. They demonstrate their method on pipes, valves, elbows, and tees for underwater pipe structures. Similarly, Guerneve~\textit{et al.}~\cite{8202215} compare sensor inputs with pre-existing CAD models, albeit by using sonar rather than laser scanners. Machado~\textit{et al.}~\cite{7783534} and dos Santos~\textit{et al.}~\cite{dos2018object} detect five classes of objects from sonar data in a harbor environment through training a neural network. Finally, McConnell~\textit{et al.}~\cite{9560737} demonstrate a method of predictive sonar-based mapping using Bayesian inference, which is able to identify and thus better reconstruct previously seen underwater objects even from partial views. 

Our method differs from the above methods in that it incorporates both camera and sonar to detect, classify, and localize \textit{previously unseen} object classes. The method extends to any object class, and does not require extensive hand-labeled training data of which there is a very little for underwater scenes. 
\section{Method}
The proposed approach consists of three primary components: Object segmentation and feature extraction; optical-acoustic fusion; and data association for mapping. Figure~\ref{fig:pipeline} shows a block diagram of the approach, illustrating how features, $3$D points, and odometry are extracted from sensor inputs and used for data association to perform semantic mapping.

\subsection{Object Segmentation and Feature Extraction}\label{subsec:segmentation}
Models leveraging unsupervised learning techniques excel in extracting visual features from unseen images~\cite{9086055}. Consequently, we select an unsupervised segmentation model MaskCut~\cite{cutler} to develop our object segmentation module.

For each image, the segmentation module outputs binary masks $\mathcal{M}_{I,i}$ of detected objects $i$ from the image $\mathcal{O}_{I}$ and simultaneously provides the masks' feature embeddings $\mathcal{F}_{I,i}$ extracted by the DINO backbone. 
The centroids $(u_{\mathcal{M},i}, v_{\mathcal{M},i})$ of each binary mask $\mathcal{M}_{I,i}$ on the image plane are extracted and used for opti-acoustic fusion, which is introduced in the following section (\ref{subsec:object_loc}).
Next, $\mathcal{F}_{I,i}$ is projected into constant dimensions $(K\times384)$, where $K \in \{1,3\}$. The original dimension of $\mathcal{F}_{I,i}$ is $(N\times384)$, with $N$ representing the number of patches for the corresponding binary mask $\mathcal{M}_{I,i}$. 
$N$, and consequently the dimension of $\mathcal{F}_{I,i}$, varies based on the size of the corresponding binary mask $\mathcal{M}_{I,i}$.

\begin{figure}
\vspace{-3mm}
\end{figure}
\begin{algorithm}[t]
\caption{Find Acoustic Range from Image Bearing}
\begin{algorithmic}[1] 
\Procedure{Bearing2Range}{$\theta$} 
\State $\theta_{s}^{\text{res}}$ = SONAR\_FOV/NUMBER\_OF\_BEAMS \tab\tab \Comment{Find angular resolution of Sonar}
\For{beam \textbf{in} beams}
\If{ $\theta_{s}$(beam) $\geq \theta - \theta_{s}^{\text{res}}/2$ 
    \textbf{and} \\  \tab\tab\tab $\theta_{s}$(beam)$\leq\theta + \theta_{s}^{\text{res}}/2$}
    \If{max(Ping($\theta_{s}$(beam))) $\geq$ $\tau$} \hfill \tab\tab\tab \hspace{1cm} \Comment{$\tau$: THRESHOLD}
    \State $Z$ = max(Ping($\theta_{s}$(beam)))
    \State \Return $Z$
    \Else{}
    \State \Return FALSE
    \EndIf
\EndIf
\EndFor
\EndProcedure
\end{algorithmic}
\label{alg:bearing2range}
\end{algorithm}
Patch Averaging (PA) and Uniform Manifold Approximation and Projection (UMAP) \cite{mcinnes2020umap} are used to accommodate masks of varying sizes while ensuring the output features maintain a constant dimension, effectively condensing the information contained in the larger-size feature embeddings into compact representation. This standardization enables a consistent fusion framework irrespective of the underlying binary mask sizes.

$\mathcal{F}_{I,i}$ is projected into feature embeddings $\mathcal{F}_{I,i}^P$ with constant dimensions $(K\times384)$, where $K \in \{1,3\}$ and $P \in \{\text{PA}, \text{UMAP}\}$.
\begin{itemize}
    \item PA: Mean feature embeddings $\mathcal{F}_{I,i}^{\text{PA}}$ of $\mathcal{F}_{I,i}$ are computed by averaging features values across the dimension $N$, yielding the size of $(1 \times 384)$.
    \item UMAP: UMAP is selected to make the feature embeddings more interpretable while preserving the global structure of $\mathcal{F}_{I,i}$ as much as possible. In this case, the size of $\mathcal{F}_{I,i}^{\text{UMAP}}$ is $(3\times384)$.
\end{itemize}




\subsection{Opti-acoustic Fusion for 3D Projection}\label{subsec:object_loc}
\begin{figure}  
    \vspace{0.5mm}
    \centering
    \scalebox{1.0}{\input{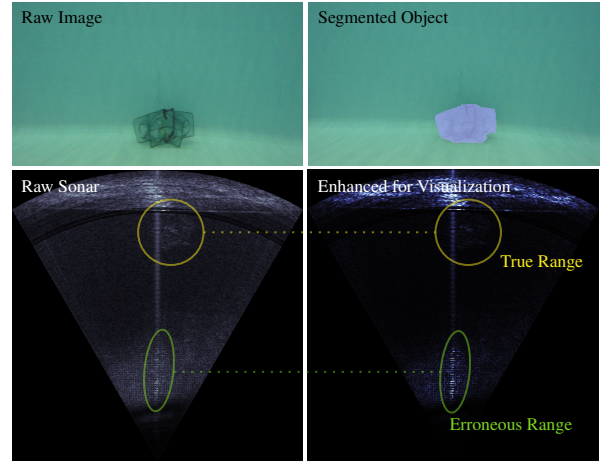}}
\caption{(Top left): Input image of a lobster cage, (Top right): Detected object, (Bottom left): Corresponding sonar frame for object range estimation, (Bottom right): Enhanced frame for visualization. Sonar data can often be noisy and is subject to multi-path effects that make the detection and range extraction of objects challenging. Our method is largely robust to such issues through the use of a probabilistic data association method that takes into account the range uncertainty that is typical with the use of sonar.}
\label{fig:sonar_range_challenge}
    \vspace{-2mm}
\end{figure}

$(u_{\mathcal{M},i}, v_{\mathcal{M},i})$ of $\mathcal{M}_{I,i}$, introduced in the previous section and camera parameters ($c_x,c_y,f_x,f_y$) are used to calculate object bearing $\theta$ and elevation $\phi_{o}$ as shown Eq~\ref{eq:theta1} and~\ref{eq:phi1}.

\begin{equation}
\theta = \arctan\left(\frac{u_{\mathcal{M},i}-c_x}{f_x}\right) = \arctan\left(\frac{X}{Z}\right)
\label{eq:theta1}
\end{equation}
\begin{equation}
\phi_o = \arctan\left(\frac{v_{\mathcal{M},i}-c_y}{f_y}\right) = \arctan\left(\frac{Y}{Z}\right)
\label{eq:phi1}
\end{equation}
Next, the acoustic beam that has the closest bearing value to $\theta$ is identified. 
 Following this selection, the range, $Z$, along the beam to the target is retrieved from a sonar sensor (Figure~\ref{fig:opti-acoustic}). 
Algorithm \ref{alg:bearing2range} provides more details on finding $Z$ from $\theta$. 
After finding $Z$, $X$ and $Y$ are calculated using the second part of Eq~\ref{eq:theta1} and~\ref{eq:phi1}. Through this procedure, the $3$D position ($X,Y,Z$) of each detected object is calculated. 
As shown in Figure~\ref{fig:sonar_range_challenge}, the range extraction method is subject to noisy and erroneous sonar range measurements due to multipath reflections. Such errors are mitigated through the use of a robust data association method, which is described in the next section. 


\subsection{Semantic SLAM}
The proposed method jointly optimizes both poses $\poses$ and landmarks $\landmarks$ given observations $\measurements$ through the \textit{maximum a posteriori} (MAP) inference problem: 

\begin{equation}
\hat{\poses}, \hat{\landmarks}=\underset{\poses \in SE(3), \landmarks \in \mathbb{R}^{3}}{\arg \max } \; p(\poses, \landmarks \mid \measurements).
\end{equation}

However, the method must also consider data association i.e. observation to landmark correspondences. Given the object class latent code as outlined in section \ref{subsec:segmentation}, and the object position as described in section \ref{subsec:object_loc}, as well as an odometry estimate (obtained as explained in Section \ref{sec:datasets}), a keyframe and landmark observation is added to the factor graph. The landmark observation is classified as belonging to a previous landmark, or as the first observation of a new landmark. This classification is determined by first checking the cosine similarity of the object observation's latent space embedding as compared to existing landmarks' embeddings. 
\begin{equation}
S_C(A, B):=\dfrac {A \cdot B} {\left\| A\right\| _{2}\left\| B\right\| _{2}} 
\end{equation}
For each existing landmark that is determined to meet a heuristically chosen similarity threshold, the Mahalanobis distance is calculated between the observed landmark and the existing landmark following the method described by Kaess~\textit{et al.}~\cite{kaess2009covariance}. The distance follows a six degree-of-freedom chi-square distribution, and thus we use the test below where $l$ is the new landmark, $l_{j}$ is the existing landmark, and $\Sigma$ is the recovered covariance matrix. 
\begin{equation}\label{eq:chi_square}
|| l - l_{j} ||_{\Sigma}^{2} <\chi_{6 dof}^2
\end{equation}
\begin{figure}[t]
    \centering
    \scalebox{1.3}{\input{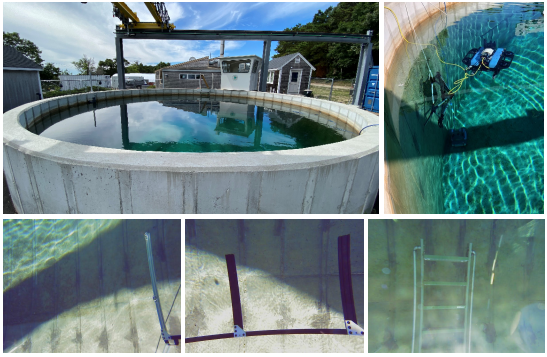}}
\caption{(Top): The outdoor tank used for data collection exhibits challenging lighting conditions, including reflections (left) and caustics (right). (Bottom): Selected raw (uncalibrated) frames from the vehicle's monocular camera. These images demonstrate the caustics (ripple lighting effect due to surface waves), reflections, and sudden lighting changes make processing underwater imagery particularly difficult compared to terrestrial scenes. Feature-based methods fail to maintain tracking due to the dynamic textures; our method uses objects instead and proves to be robust to such lighting effects.}
\label{fig:lighting_challenges}
\end{figure}
If no landmarks meet both criteria above, then a new landmark is added. Otherwise, the max-likelihood hypothesis is selected given prior estimates $\poses^0, \landmarks^0$. 
\begin{equation}
\hat{\associations}=\underset{\associations}{\arg \max } \; p\left(\associations \mid \poses^0, \landmarks^0, \measurements\right)
\end{equation}
where $\associations \triangleq \{\association_k : \association_k \in \mathbb{N}_{\leq M},\ k = 1, \ldots K\}$ and $M$ is the number of existing landmarks. 
Given the selected data association, the pose and landmark estimates are then updated: 
\begin{equation}
\hat{\poses}, \hat{\landmarks}=\underset{\poses \in SE(3), \landmarks \in \mathbb{R}^{3}}{\arg \max } \; \log p(\measurements \mid \poses, \landmarks, \hat{\associations}).
\end{equation}
By utilizing a factor graph framework and iSAM2 \cite{kaess2012isam2}, it is possible to efficiently and accurately solve the above inference problems.

\section{Datasets}\label{sec:datasets}
Two different settings with several different configurations were used to collect datasets with a variety of sensors, lighting conditions, and objects to test the proposed method.





\setlength{\fboxrule}{0pt}
\begin{figure}[t]
\vspace{1mm}
\fbox{\begin{subfigure}[b]{\linewidth}
        \centering
  \includegraphics[scale=0.39,trim={0.5cm 0.8cm 1.1cm 2.3cm},clip]{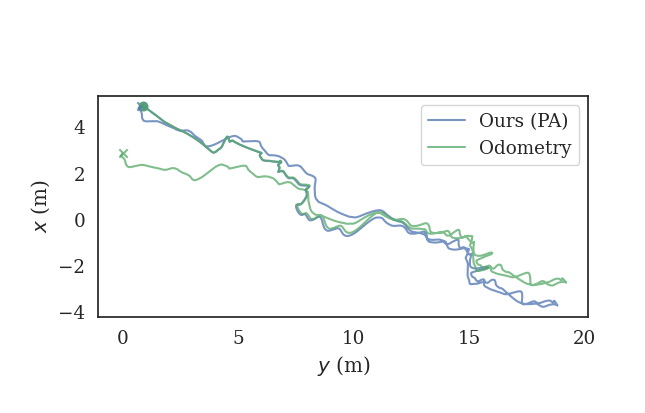}
  \caption{Run 3: 2 Laps}
  \end{subfigure}}

\fbox{\begin{subfigure}[b]{\linewidth}
        \centering
  \includegraphics[scale=0.29,trim={0.5cm 0.5cm 0.5cm 0.3cm},clip]{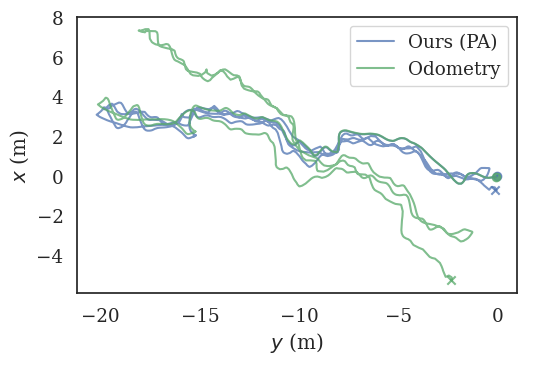}
  \caption{Run 4: 4 Laps}
  \end{subfigure}}
\caption{Comparison results of trajectory estimates, where the 'O' and 'X' markers denote the start and end points of each lap, respectively. The start and end points of the true trajectory are the same location, but drift causes the vehicle's odometric estimate to deviate over time. (a) The two lap trajectory (Run 3) estimated by our patch averaging method (PA) can reduce drift errors from the odometry trajectory. (b) A similar comparison over four laps (Run 4), highlighting corrections made by our method against the odometry. Motion in the z-direction is negligible and thus is not shown.}

\label{fig:traj_comparisons}
\end{figure}
\begin{figure*}[h]
    \vspace{1mm}
\center
    \scalebox{1.2}{\input{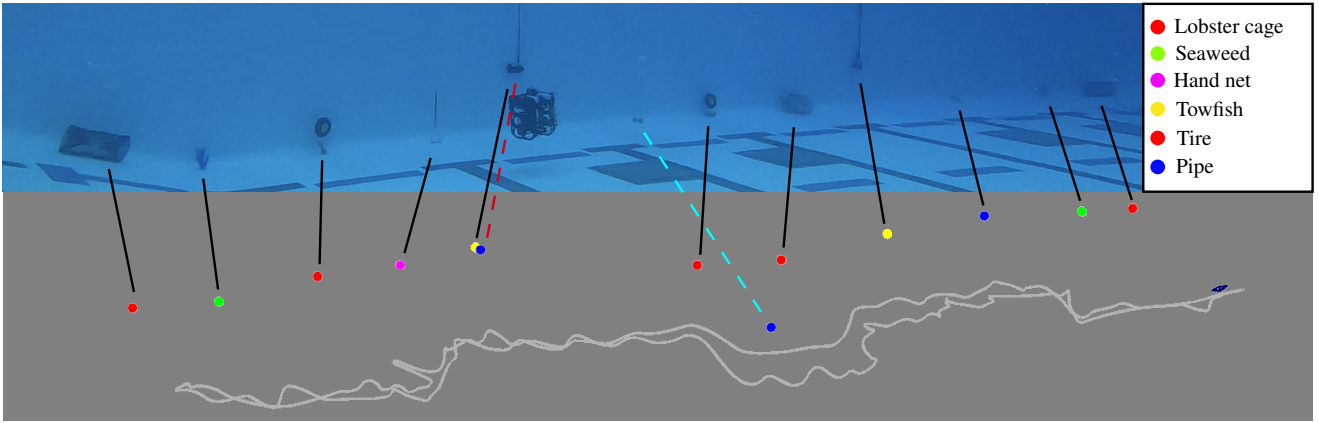}}
  \caption{The upper portion of the image illustrates an underwater vehicle navigating through an underwater environment where various objects are placed.
  The trajectory map below shows the estimated trajectory of the vehicle and the location of objects encountered during the mission. The colors are assigned based on semantic class groupings. Note that the tire and lobster cage are unexpectedly similar in the latent space and thus appear as the same class. Black lines represent correct matches, a red dashed line shows false positives due to an error from feature embedding, and a cyan dashed line is a misaligned object match in the map due to an error from range values from the sonar sensor.}

  \label{fig:map_with_legend}
  \vspace{-4mm}
\end{figure*}  





\subsection{Indoor Pool}

Data was collected in an indoor tank with a large variety of objects (Run 1-4 in Table~\ref{tab:indoor_objects}), but no available ground truth trajectory. The vehicle used for these experiments is a modified BlueROV2, which is described in detail in \cite{Osedach2020}. 
Odometry is obtained by fusing a Teledyne Impulse DVL, a VectorNav IMU, and a Bar30 pressure sensor using an Extended Kalman Filter. The camera is a calibrated Sony Exmor IMX322 low-light camera with an 80-degree horizontal field-of-view and a 64-degree vertical field-of-view. The sonar is a Oculus m1200d multibeam operating at 2.1MHz, which provides a 2.5 mm range resolution, a 0.4 degree angular resolution, a 60 degree horizontal aperture, and a 12 degree vertical aperture. Due to the vertical aperture, camera data is needed to constrain the objects' elevation angles.  

\subsection{Outdoor Tank}
Data was also collected in an outdoor tank on a sunny day with a modified BlueROV2 that has an acoustic array for the iLBL-USBL measurements attached to the top of the vehicle. The vehicle also has a Bar30 pressure sensor for depth measurements, the same camera as in the indoor datasets, and an Oculus m750d multibeam sonar operating at 750kHz frequency with a 5 Hz update rate. In this mode, the sonar has a range resolution of 4 mm, an angular resolution of 1 degree, a horizontal field-of-view of 130 degrees, and a vertical aperture of 20 degrees. The relatively large vertical aperture, which cannot be resolved in the received data, means that the use of the camera for estimating the elevation angle of objects is crucial. 

The acoustic beacons are used to provide odometry estimates, and are also free of the magnetic interference that prevented the use of magnetometer-based heading measurements. 
One-way travel time range measurements from each of the two beacons to the vehicle's receiver array create a nonlinear least-squares optimization as described below to uniquely determine the vehicle position from overlapping range circles.

\begin{align}
(\tilde{x},\tilde{y}) =& 
\underset{x, y}{\operatorname{min}} \sum_{k=1}^M [\sqrt{(x-x_{k})^2+(y-y_{k})^2}-r_{k}]^2
\end{align}
where $(x_k, y_k)$ is the position of beacon $k$, and $r_k$ is the range to beacon $k$.

Since magnetometer-based heading measurements are adversely affected by the magnetic interference of the tank walls and other nearby equipment, to get a more accurate ground truth trajectory, the vehicle heading is directly inferred from beamforming on the measurements from the vehicle's acoustic array by solving for vehicle heading $\theta_k$:
\begin{align}
\theta_k  &= [\phi_k - \arctan2(\tilde{y}-y_k, \tilde{x}-x_k) - \frac{\pi}{2}] \mod 2\pi. \label{eq:acoheading}
\end{align}

\newcolumntype{M}[1]{>{\centering\arraybackslash}m{#1}} 

\begin{table}[]
\centering
\caption{Summary of Datasets}
\label{tab:indoor_objects}
\begin{tabularx}{\linewidth}{M{0.4cm} X c}
\hline
\textbf{Run} & \multicolumn{1}{c}{\textbf{Objects}}                                                                                   & \textbf{Laps} \\ \hline
1            & $1$ small hand net, $1$ large hand net, $2$ sets of stairs, $2$ recycling bins, $2$ types of towfish, $2$ trash bins, $2$ paddles       & $2$             \\ \hline
2            & $1$ model submarine, $1$ green seaweed patch, $1$ pipe, $1$ towfish, $1$ star-shaped lobster cage, $1$ tire, $1$ small hand net        & $2$             \\ \hline
3            & $2$ round lobster cages, $1$ green seaweed patch, $1$ blue seaweed patch, $2$ tires, $1$ small hand net, $2$ types of towfish, $2$ pipes with anchors, $1$ star-shaped lobster cage & $2$             \\ \hline
4            & $2$ round lobster cages, $1$ green seaweed patch, $1$ blue seaweed patch, $2$ tires, $1$ small hand net, $2$ types of towfish, $2$ pipes with anchors, $1$ star-shaped lobster cage & $4$             \\ \hline
5            & $1$ fish cage, $1$ ladder, $1$ pipe, $1$ T-slot extrusion structure & $3$             \\ \hline
\end{tabularx}
\end{table}

The vehicle position and heading are available at $1$ Hz, and are directly used as the absolute (as opposed to relative as in the indoor datasets) odometry input for the semantic SLAM method. The ground truth trajectory is further refined through a pose-graph SLAM method that incorporates ICP-based alignments of the wall and structures. 
The use of an outdoor tank on a sunny day means that the data has several difficult lighting effects, as seen in Figure \ref{fig:lighting_challenges}.  
The dataset consists of a single run (Run 5 in Table~\ref{tab:indoor_objects}) of three loops around the circular tank, which is roughly 64 square meters in area with an average depth of 4 meters. Note that while certain markings on the wall due to algae build-up were detected by the open-set system, those objects are not included the map accuracy results in Table~\ref{table:precision_recall}.

\section{Results}

The proposed method is thoroughly tested on a variety of collected datasets, with map accuracy results shown in Table~\ref{table:precision_recall}, and trajectory accuracy along with an ablation study in Table~\ref{table:whoi_ablation}. Quantitative trajectory evaluations are available for the outdoor tank, but qualitative improvements of the trajectory can be seen in the indoor pool data as well in Figure~\ref{fig:traj_comparisons}. While the start and end point should be the same, the effects of odometric drift can be clearly seen as the end points of the odometry only trajectories are several meters from the start point. In contrast, our method is able to make loop closures through successfully re-identifying and associating objects as the vehicle returns to the start point, thus allowing the tightly-coupled optimization to correct the odometric drift. 

The outdoor tank dataset demonstrates the failure of feature-based methods in such conditions, as the standard implementation for such methods, ORB-SLAM3~\cite{ORBSLAM3_TRO} is unable to complete more than $27\%$ of the trajectory due to the dynamic movement of caustics throughout the trajectory. Furthermore, our method is able to improve the trajectory accuracy from the odometry through identifying previously seen objects. The results also emphasize the success of the use of the semantic encoding of each object (\textit{e.g.,} Figure~\ref{fig:map_with_legend}), as the ablation demonstrates that the trajectory is made worse without the assistance of the semantic latent codes due to incorrect data associations. 

\begin{table}[t]
    \vspace{2mm}
\begin{center}
\caption{Map accuracy}
\begin{tabular}{c c c c } 
 \hline
 Dataset & Method & Precision & Recall\\ [0.1ex] 
 \hline 
  & Patch Averaging (ours) & 0.82 & 0.82 \\ 
  Indoor, Run 1 & UMAP (ours) & 0.58 & 0.78 \\
  & YOLOv8 & 1.0 & 1.0 \\ 
 \hline
  & Patch Averaging (ours) & 0.41 & 1.0 \\ 
  Indoor, Run 2 & UMAP (ours) & 0.38 & 0.71 \\
  & YOLOv8 & 1.0 & 0.75 \\ 
 \hline
  & Patch Averaging (ours) & 0.86 & 1.0 \\ 
  Indoor, Run 3 & UMAP (ours) & 0.56 & 0.82 \\
  & YOLOv8 & 1.0 & 0.83 \\ 
 \hline
  & Patch Averaging (ours) & 0.73 & 0.92 \\ 
  Indoor, Run 4 & UMAP (ours) & 0.48 & 0.92 \\
  & YOLOv8 & 1.0 & 0.83 \\ 
 \hline

& Patch Averaging (ours) & 0.75 & 0.60 \\ 
  Outdoor Tank, Run 5 & UMAP (ours) & 0.66 & 0.75 \\
  & YOLOv8 & 1.0 & 1.0 \\ 
 \hline
\end{tabular}
\label{table:precision_recall}
\end{center}
\end{table}

The proposed method also achieves strong map accuracy. PA outperforms UMAP consistently, which is surprising since patch averaging only allows for a 384 length encoding, whereas UMAP effectively has a length 1,152 encoding, which should allow for more expressiveness. 
However, this unexpected outcome is likely attributed to a fundamental discrepancy in the underlying assumptions about data distribution. UMAP's effectiveness is based on the premise that data is uniformly distributed across a manifold, an assumption that may not hold for the features extracted in our pipeline.

A baseline method based on fine-tuning YOLOv8 is also implemented. While off-the-shelf network weights do not identify any of the objects in the collected underwater datasets, 230 hand-labeled images including every object class in the datasets were used to fine-tune a YOLOv8n network over 25 epochs. The 230 images were augmented with rotation and blurring to produce a dataset of 691 training images. The labeled images are included as part of the dataset release. Object detections are encoded as one-hot vectors, with the rest of the system pipeline the same as our method. While the baseline method performs well, it has received training data of the exact objects in the same scene and lighting conditions as the test data, unlike our method which is able to identify and classify objects while receiving no labeled training data or prior knowledge of the scene. 

It should be noted that the precision and recall metrics are at the \textit{map} level, not observation level. Thus, each correct object identification requires the correct data association over all observations of the object, which can be over 100 individual observations. However, just one bad data association decision can cause the creation of a new object, which then adversely affects the precision figures. False positives from incorrect data associations occur for two different reasons: 1) very incorrect ranges due to multi-path effects in the sonar data as seen in Figure \ref{fig:sonar_range_challenge}, and 2) difficulty in determining that an observation is of the same class as an existing landmark due to light changes (Figure \ref{fig:lighting_challenges}) or viewing angle changes. False negatives occur solely due to the segmentation network, as it often is unable to detect the white pipe object as there is very little contrast with the background. Nevertheless, the overall map accuracy results demonstrate that our system is largely robust to such issues, and is able to achieve comparable results to a closed-set fine-tuned baseline system.

\begin{table}[t]
    \vspace{2mm}
\begin{center}
\caption{Outdoor Dataset, Average Pose Error in meters. Note that the ORBSLAM3 Baseline method loses tracking and fails to complete trajectory. The error metric for ORBSLAM3 is shown for completed section only (27\% of full trajectory).}
\begin{tabular}{c c } 
 \hline
 Method & APE (m) \\ [0.1ex] 
 \hline 
 Odometry & 0.58 \\ 
 Odom + Geometric & 1.40 \\
 Odom + Geo + Uncertainty & 0.60 \\
 Odom + Geo + Uncert + Semantics & \textbf{0.56} \\
 \hline 
 Baseline: ORBSLAM3 & 2.28 \\
 Baseline: Closed-Set (YOLOv8) & 0.59 \\
 \hline
\end{tabular}

\label{table:whoi_ablation}
\end{center}
\vspace{-2mm}
\end{table}

\section{Conclusion}
A method that combines monocular camera data with sonar data for identifying, localizing, and mapping underwater objects is proposed. A robust data association method is implemented along with a SLAM system using a factor graph representation to jointly optimize landmark and vehicle poses. A variety of datasets in challenging conditions unique to underwater scenarios with many different typical underwater objects were collected and open-sourced, and the method is evaluated on these datasets. The method performs well quantitatively and qualitatively on mapping accuracy as well as trajectory accuracy as compared to baseline methods.

\section*{Acknowledgments}
\footnotesize 
The authors would like to thank Thomas Jeongho Song, John Paul Morrison, and Ben Weiss for their assistance in data collection. 

This work was supported by the MIT Lincoln Laboratory Autonomous Systems Line which is funded by the Under Secretary of Defense for Research and Engineering through Air Force Contract No. FA8702-15-D-0001, ONR grants
N00014-18-1-2832, N00014-23-12164, and N00014-19-1-2571 (Neuroautonomy MURI), and the MIT Portugal Program.

\bibliographystyle{IEEEtran}
\bibliography{main}

\end{document}